\documentclass[runningheads]{llncs}
\usepackage{graphicx}
\usepackage{comment}
\usepackage{amsmath,amssymb} 
\usepackage{color}
\usepackage{microtype}

\usepackage{subfig}
\usepackage{multirow}
\usepackage{makecell}
\usepackage{caption}
\usepackage[ruled,vlined]{algorithm2e}
\usepackage{placeins}
\usepackage{dblfloatfix}
\usepackage{hyperref}
\hypersetup{
  colorlinks = true, 
  urlcolor   = magenta, 
  linkcolor  = blue, 
  citecolor  = red 
}

\usepackage[dvipsnames]{xcolor}

\newcommand{\addition}[1]{#1}

\newcommand{\mytilde}{{\raise.17ex\hbox{$\scriptstyle\sim$}}}
\newcolumntype{?}{!{\vrule width 1pt}}

\begin{document}
\pagestyle{headings}
\mainmatter
\def\ECCVSubNumber{4458}  

\title{Active Perception using Light Curtains for Autonomous Driving}

\titlerunning{Active Perception using Light Curtains for Autonomous Driving}
%
\author{Siddharth Ancha \and
Yaadhav Raaj \and
Peiyun Hu \and\\
Srinivasa G. Narasimhan \and
David Held}
\authorrunning{S. Ancha et al.}
%
\institute{Carnegie Mellon University, Pittsburgh PA 15213, USA\\
\email{\{sancha,ryaadhav,peiyunh,srinivas,dheld\}@andrew.cmu.edu}}
\maketitle

\begin{footnotesize}
    \hspace{-10pt}
    Website: \fontsize{8}{8}\href{http://siddancha.github.io/projects/active-perception-light-curtains}{\texttt{http://siddancha.github.io/projects/active-perception-light-curtains}}
\end{footnotesize}

\vspace{-1pt}

\begin{abstract}
Most real-world 3D sensors such as LiDARs \addition{perform fixed scans of the entire environment}, while being decoupled from the recognition system that processes the sensor data. In this work, we propose a method for 3D object recognition using light curtains, a resource-efficient \addition{\textit{controllable}} sensor that measures depth at user-specified locations in the environment. Crucially, we propose using prediction uncertainty of a deep learning based 3D point cloud detector to guide \addition{active perception.} Given a neural network's uncertainty, we derive an optimization objective to place light curtains using the principle of maximizing information gain. Then, we develop a novel and efficient optimization algorithm
to maximize this objective by encoding the physical constraints of the device into a constraint graph and optimizing with dynamic programming. We show how a 3D detector can be trained to detect objects in a scene by sequentially placing uncertainty-guided light curtains to successively improve detection accuracy. Links to code can be found on the project webpage.

\end{abstract}

\newcommand{\X}{\mathbf{X}}
\newcommand{\R}{\mathbf{R}}
\newcommand{\D}{\mathcal{D}}
\newcommand{\dtmax}{\Delta \theta_\text{max}}

\section{Introduction}

3D sensors, such as LiDAR, have become ubiquitous for perception in autonomous systems operating in the real world, such as self-driving vehicles and field robots. Combined with recent advances in deep-learning based visual recognition systems, they have lead to significant breakthroughs in perception for autonomous driving, enabling the recent surge of commercial interest in self-driving technology.

However, most 3D sensors in use today \addition{perform \textit{passive perception}}, meaning that they continuously sense the entire environment while being completely decoupled from the recognition system that will eventually process the sensor data. In such a case, sensing the entire scene can be potentially inefficient. For example, consider an object detector running on a self-driving car that is trying to recognize objects in its environment. Suppose that it is confident that a tree-like structure on the side of the street is not a vehicle, but it is unsure whether an object turning around the curb is a vehicle or a pedestrian. In such a scenario, it might be beneficial if the 3D sensor focuses on collecting more data from the latter object, rather than distributing its sensing capacity uniformly throughout the scene.

\begin{figure}
  \centering
  \includegraphics[width=\textwidth]{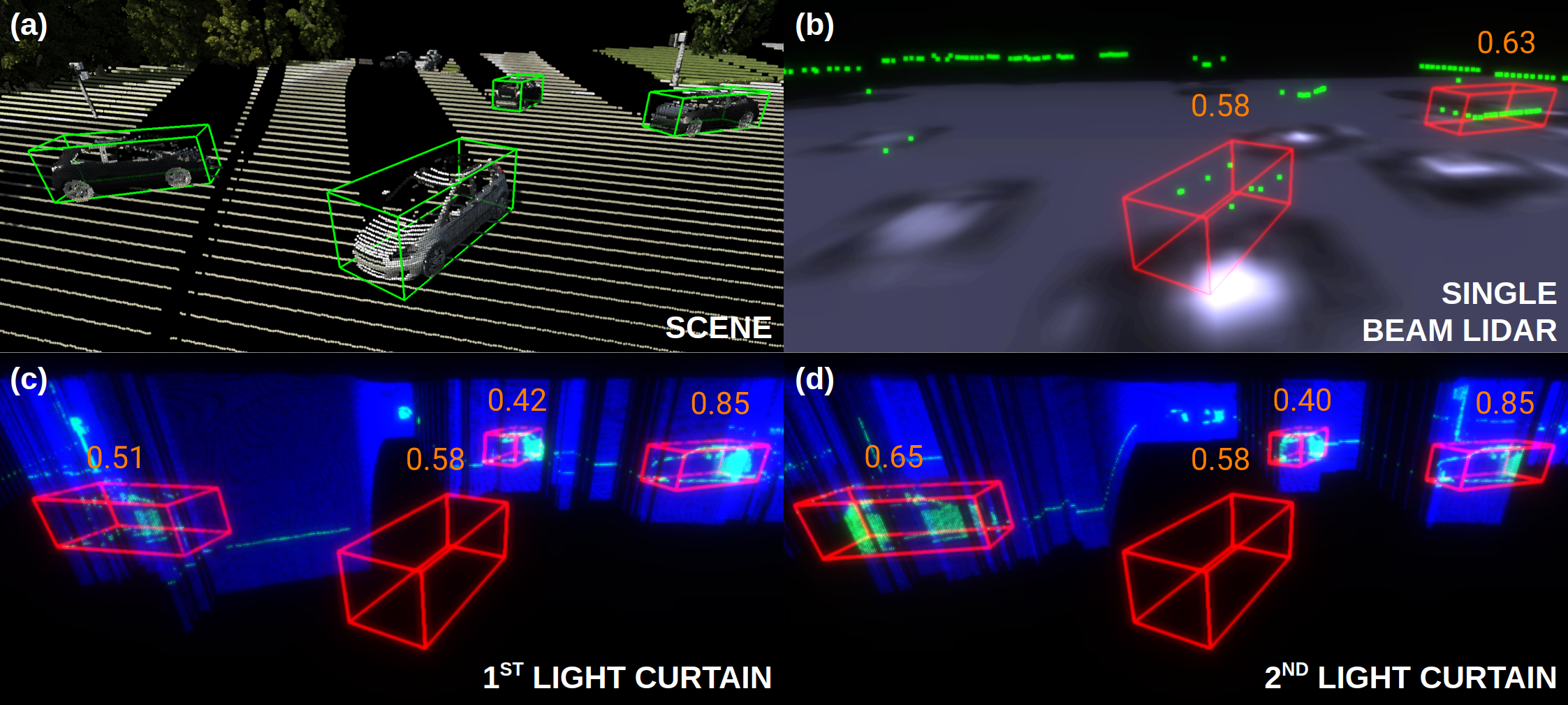}
  \caption{\textit{Object detection using light curtains.} (a) Scene with 4 cars; ground-truth boxes shown in green. (b) Sparse green points are from a single-beam LiDAR; it can detect only two cars (red boxes). Numbers above detections boxes are confidence scores. Uncertainty map in greyscale is displayed underneath: whiter means higher uncertainty. (c) First light curtain (blue) is placed to optimally cover the most uncertain regions. Dense points (green) from light curtain results in detecting 2 more cars. (d) Second light curtain senses even more points and fixes the misalignment error in the leftmost detection.}
  \label{fig:pull}
  \vspace{-10pt}
\end{figure}

In this work, we propose a method for 3D object detection \addition{using sensors that perform \textit{active perception}}, i.e. sensors that can be purposefully controlled to sense specific regions in the environment. Programmable light curtains~\cite{wang2018programmable,Bartels_2019_ICCV} were recently proposed as \addition{controllable}, light-weight, and resource efficient sensors that measure the presence of objects intersecting any vertical ruled surface whose shape can be specified by the user (see Fig.~\ref{fig:illustration}). There are two main advantages of using programmable light curtains over LiDARs. First, they can be cheaply constructed, since light curtains use ordinary CMOS sensors (a current lab-built prototype costs \$1000, and the price is expected to go down significantly in production). In contrast, a 64-beam Velodyne LiDAR that is commonly used in 3D self-driving datasets like KITTI~\cite{geiger2013vision} costs upwards of \$80,000. Second, light curtains generate data with much higher resolution in regions where they actively focus~\cite{Bartels_2019_ICCV} while LiDARs sense the entire environment and have low spatial and angular resolution.

One weakness of light curtains is that they are able to sense only a subset of the environment -- a vertical ruled surface (see Fig.~\ref{fig:pull}(c,d), Fig~\ref{fig:illustration}). In contrast, a LiDAR senses the entire scene.  To mitigate this weakness, we can take advantage of the fact that the light curtain is a \addition{\emph{controllable}} sensor -- we can choose where to place the light curtains. Thus, we must intelligently place light curtains in the appropriate locations, so that they sense the most important parts of the scene. In this work, we develop an algorithm for determining how to best place the light curtains for maximal detection performance.

We propose to use a deep neural network's prediction uncertainty as a guide for determining how to actively sense an environment. Our insight is that if a \addition{controllable} sensor images the regions which the network is most uncertain about, the data obtained from those regions can help resolve the network's uncertainty and improve recognition. Conveniently, most deep learning based recognition systems output confidence maps, which can be used for this purpose when converted to an appropriate notion of uncertainty.

Given neural network uncertainty estimates, we show how a light curtain can be  placed to \textit{optimally} cover the regions of maximum uncertainty. First, we use an information-gain based framework to propose placing light curtains that maximize the sum of uncertainties of the covered region (Sec.~\ref{sec:objective}, Appendix~\ref{app:info-gain}). However, the structure of the light curtain and physical constraints of the device impose restrictions on how the light curtain can be placed. Our novel solution is to precompute a ``constraint graph'', which describes all possible light curtain placements that respect these physical constraints. We then use an optimization approach based on dynamic programming to efficiently search over all possible feasible paths in the constraint graph and maximize this objective (Sec.~\ref{sec:dp}). This is a novel approach to constrained optimization of a controllable sensor's trajectory which takes advantage of the properties of the problem we are trying to solve.


Our proposed active \addition{perception} pipeline for 3D detection proceeds as follows. We initially record sparse data with an inexpensive single beam LIDAR sensor \addition{that performs fixed 3D scans}.  This data is input to a 3D point cloud object detector, which outputs an initial set of detections and confidence estimates.  These confidence estimates are converted into uncertainty estimates, which are used by our dynamic programming algorithm to determine where to place the first light curtain.  The output of the light curtain readings are again input to the 3D object detector to obtain refined detections and an updated uncertainty map. This process of estimating detections and placing new light curtains can be repeated multiple times (Fig.~\ref{fig:pipeline}). Hence, we are able to sense the environment progressively, intelligently, and efficiently.

We evaluate our algorithm using two synthetic datasets of urban driving scenes~\cite{vkitti,synthia}.  Our experiments demonstrate that our algorithm leads to a monotonic improvement in performance with successive light curtain placements.
We compare our proposed optimal light curtain placement strategy to multiple baseline strategies and find that they are significantly outperformed by our method.
To summarize, our contributions are the following:
\begin{itemize}
  \item[$\bullet$] We propose a method for using a deep learning based 3D object detector's prediction uncertainty as a guide for active sensing \addition{(Sec.~\ref{sec:uncertainty-extraction})}.
  \item[$\bullet$] \addition{Given a network's uncertainty, we derive an optimization objective to decide where to place light curtains using the principle of maximizing information gain (Sec.~\ref{sec:objective}, Appendix~\ref{app:info-gain}).}
  \item[$\bullet$]
  Our novel contribution is to encode the physical constraints of the device into a graph and use dynamic-programming based graph optimization to efficiently maximize the objective while satisfying the physical constraints (Sec.~\ref{sec:objective}, \ref{sec:dp}).
  \item[$\bullet$] We show how to train such an active detector using online light curtain data generation \addition{(Sec.~\ref{sec:online-training})}.
  \item[$\bullet$] We empirically demonstrate that our approach \addition{successively improves detection performance over LiDAR and} is significantly better compared to a number of baseline approaches \addition{(Sec.~\ref{sec:experiments})}.
\end{itemize}

\section{Related Work}
\label{sec:related-work}


\subsection{Active Perception}
Active Perception encompasses a variety of problems and techniques that involve actively controlling the sensor for improved perception~\cite{bajcsy1988active,wilkes1994active}. Examples include actively modifying camera parameters~\cite{bajcsy1988active}, moving a camera to look around occluding objects~\cite{cheng2018reinforcement}, and obtaining the next-best-view~\cite{connolly1985determination}. Prior works have used active perception for static scenes~\cite{mnih2014recurrent,ba2014multiple} via a series of controllable partial glimpses. 
Our paper differs from past work because we use a controllable depth sensor (light curtains) and combine it with deep learning uncertainty estimates in a novel active perception algorithm.

\subsection{Object Detection from Point Clouds}
There have been many recent advances in deep learning  for 3D object detection. Approaches include representing LiDAR data as range images in LaserNet\cite{meyer2019lasernet}, using raw point clouds~\cite{shi2019pointrcnn}, and using point clouds in the bird's eye view such as AVOD~\cite{ku2018joint}, HDNet~\cite{yang2018hdnet} and Complex-YOLO~\cite{simony2018complex}. Most state-of-the-art approaches use voxelized point clouds, such as VoxelNet~\cite{zhou2018voxelnet}, PointPillars~\cite{lang2019pointpillars}, SECOND~\cite{yan2018second}, and CBGS~\cite{zhu2019class}. These methods process an input point cloud by dividing the space into 3D regions (voxels or pillars) and extracting features from each of region using a PointNet~\cite{qi2017pointnet} based architecture. Then, the volumetric feature map is converted to 2D features via convolutions, followed by a detection head that produces bounding boxes.
We demonstrate that we can use such detectors, along with our novel light curtain placement algorithm, to process data from a single beam LiDAR combined with light curtains.

\subsection{Next-Best View Planning}
\label{sec:related-work-nbv}
Next-best view (NBV) planning refers to a broad set of problems in which the objective is to select the next best sensing action in order to solve a specific task. Typical problems include object instance classification~\cite{Wu_2015_CVPR,doumanoglou2016recovering,denzler2002information,scott2003view} and 3D reconstruction~\cite{isler2016information,kriegel2015efficient,vasquez2014volumetric,daudelin2017adaptable,haner2012covariance}. Many works on next-best view formulate the objective as maximizing information gain (also known as mutual information)~\cite{Wu_2015_CVPR,denzler2002information,isler2016information,kriegel2015efficient,vasquez2014volumetric,daudelin2017adaptable}, using models such as probabilistic occupancy grids for beliefs over states~\cite{Wu_2015_CVPR,isler2016information,kriegel2015efficient,vasquez2014volumetric,daudelin2017adaptable}.
Our method is similar in spirit to next-best view. One could consider each light curtain placement as obtaining a new view of the environment; we try to find the next best light curtain that aids object detection. In Sec.~\ref{sec:objective} and Appendix~\ref{app:info-gain}, we derive an information-gain based objective to find the next best light curtain placement.

\section{Background on Light Curtains}
\label{sec:background}
\begin{figure}[t]
    \vspace{-15pt}
    \centering
    \subfloat[Working principle]{
        \includegraphics[trim=0 0 0 0,clip,width=0.35\textwidth]{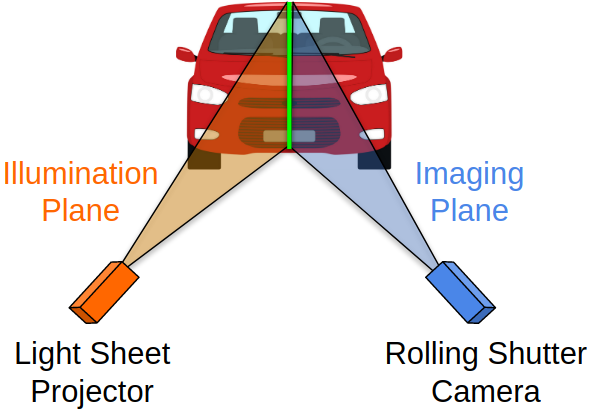}
    }
    \subfloat[Optical schematic (top view)]{
        \includegraphics[trim=0 0 0 0,clip,width=0.45\textwidth]{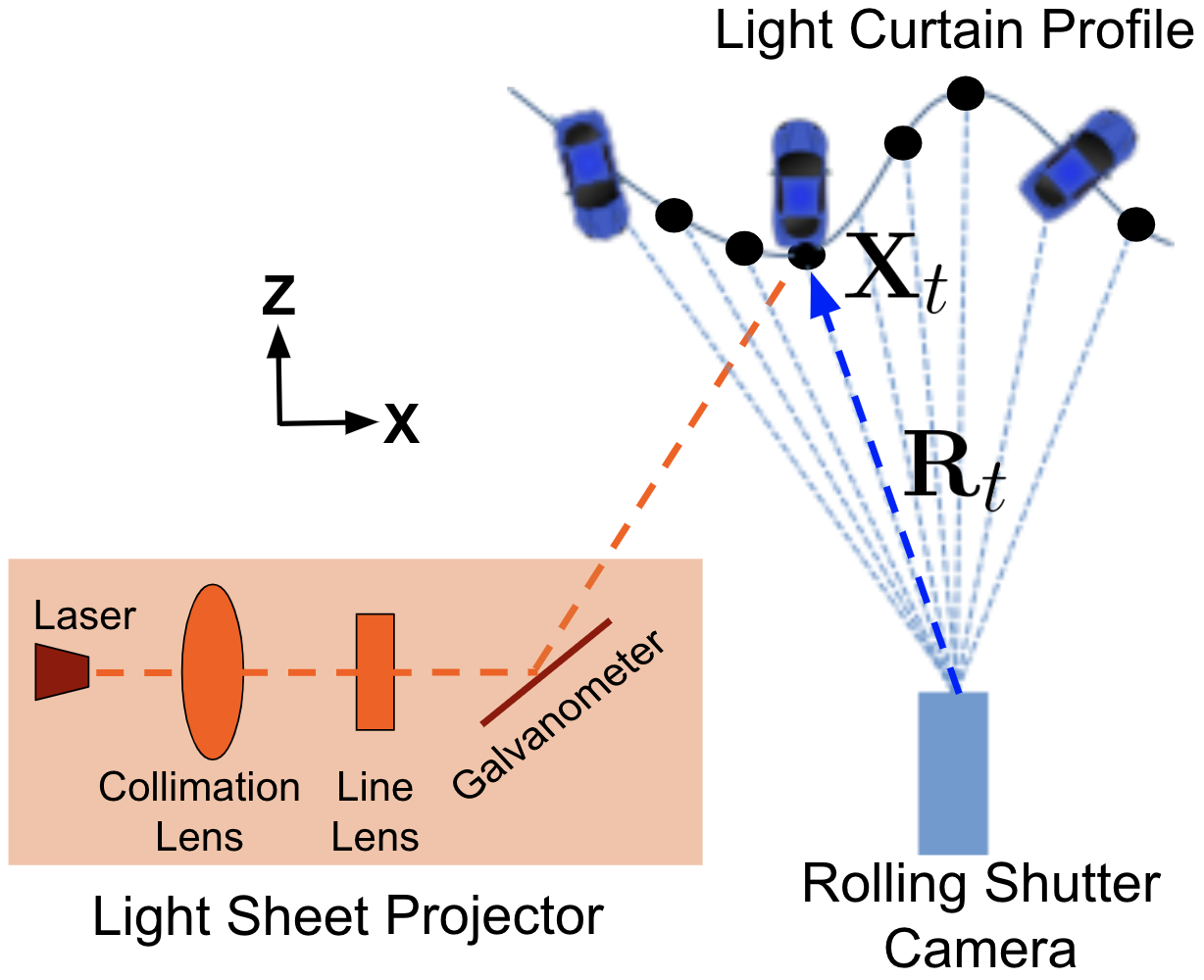}
    }
    \caption{Illustration of programmable light curtains adapted from~\cite{Bartels_2019_ICCV,wang2018programmable}. a) The light curtain is placed at the intersection of the illumination plane (from the projector) and the imaging plane (from the camera). b) A programmable galvanometer and a rolling shutter camera create multiple points of intersection, $\X_t$.
    }
    \label{fig:illustration}
    \vspace{-10pt}
\end{figure}


\noindent Programmable \textit{light curtains}~\cite{wang2018programmable,Bartels_2019_ICCV} are a sensor for adaptive depth sensing. ``Light curtains'' can be thought of as virtual surfaces placed in the environment. They can detect points on objects that intersect this surface. Before explaining how the curtain is created, we briefly describe our coordinate system and the basics of a rolling shutter camera.\\
\textbf{Coordinate system:} Throughout the paper, we will use the standard camera coordinate system centered at the sensor. We assume that the $z$ axis corresponds to depth from the sensor pointing forward, and that the $y$ vector points vertically downwards. Hence the $xz$-plane is parallel to the ground and corresponds to a top-down view, also referred to as the bird's eye view.\\
\textbf{Rolling shutter camera}: A rolling shutter camera contains pixels arranged in $T$ number of vertical columns. Each pixel column corresponds to a vertical imaging plane. Readings from only those visible 3D points that lie on the imaging plane get recorded onto its pixel column. We will denote the $xz$-projection of the imaging plane corresponding to the $t$-th pixel column by ray $\R_t$, shown in the top-down view in Fig.~\ref{fig:illustration}(b). We will refer to these as ``camera rays''. The camera has a rolling shutter that successively activates each pixel column and its imaging plane one at a time from left to right. The time interval between the activation of two adjacent pixel columns is determined by the pixel clock.\\
\textbf{Working principle of light curtains:} The latest version of light curtains~\cite{Bartels_2019_ICCV} works by rapidly rotating a light sheet laser in synchrony with the motion of a camera's rolling shutter. A laser beam is collimated and shaped into a line sheet using appropriate lenses and is reflected at a desired angle using a controllable galvanometer mirror (see Fig.~\ref{fig:illustration}(b)). The illumination plane created by the laser intersects the active imaging plane of the camera in a vertical line along the curtain profile (Fig.~\ref{fig:illustration}(a)). The $xz$-projection of this vertical line intersecting the $t$-th imaging plane lies on $\R_t$, and we call this the $t$-th ``control point", denoted by $\X_t$ (Fig.~\ref{fig:illustration}(b)).\\
\textbf{Light curtain input}: The shape of a light curtain is uniquely defined by where it intersects each camera ray in the $xz$-plane, i.e. the control points $\{\X_1, \dots, \X_T\}$. These will act as inputs to the light curtain device. In order to produce the light curtain defined by $\{\X_t\}_{t=1}^T$, the galvanometer is programmed to compute and rotate at, for each camera ray $\R_t$, the reflection angle $\theta_t(\X_t)$ of the laser beam such that the laser sheet intersects $\R_t$ at $\X_t$. By selecting a control point on each camera ray, the light curtain device can be made to image any vertical ruled surface~\cite{Bartels_2019_ICCV,wang2018programmable}.\\
\textbf{Light curtain output}: The light curtain outputs a point cloud of all 3D visible points in the scene that intersect the light curtain surface. The density of light curtain points on the surface is usually much higher than LiDAR points.\\
\textbf{Light curtain constraints}: The rotating galvanometer can only operate at a maximum angular velocity $\omega_\text{max}$.
Let $\X_t$ and $\X_{t+1}$ be the control points on two consecutive camera rays $\R_t$ and $\R_{t+1}$. These induce laser angles $\theta(\X_t)$ and $\theta(\X_{t+1})$ respectively. If $\Delta t$ is the time difference between when the $t$-th and $(t+1)$-th pixel columns are active, the galvanometer needs to rotate by an angle of $\Delta \theta(\X_t) = \theta(\X_{t+1}) - \theta(\X_t)$ within $\Delta t$ time. Denote $\dtmax = \omega_\text{max} \cdot \Delta t$. Then the light curtain can only image control points subject to $|\theta(\X_{t+1}) - \theta(\X_t)| \leq \dtmax,\ \forall 1 \leq t < T$.


\section{Approach}
\subsection{Overview}
\begin{figure}[t]
    \centering
    \includegraphics[width=0.95\textwidth]{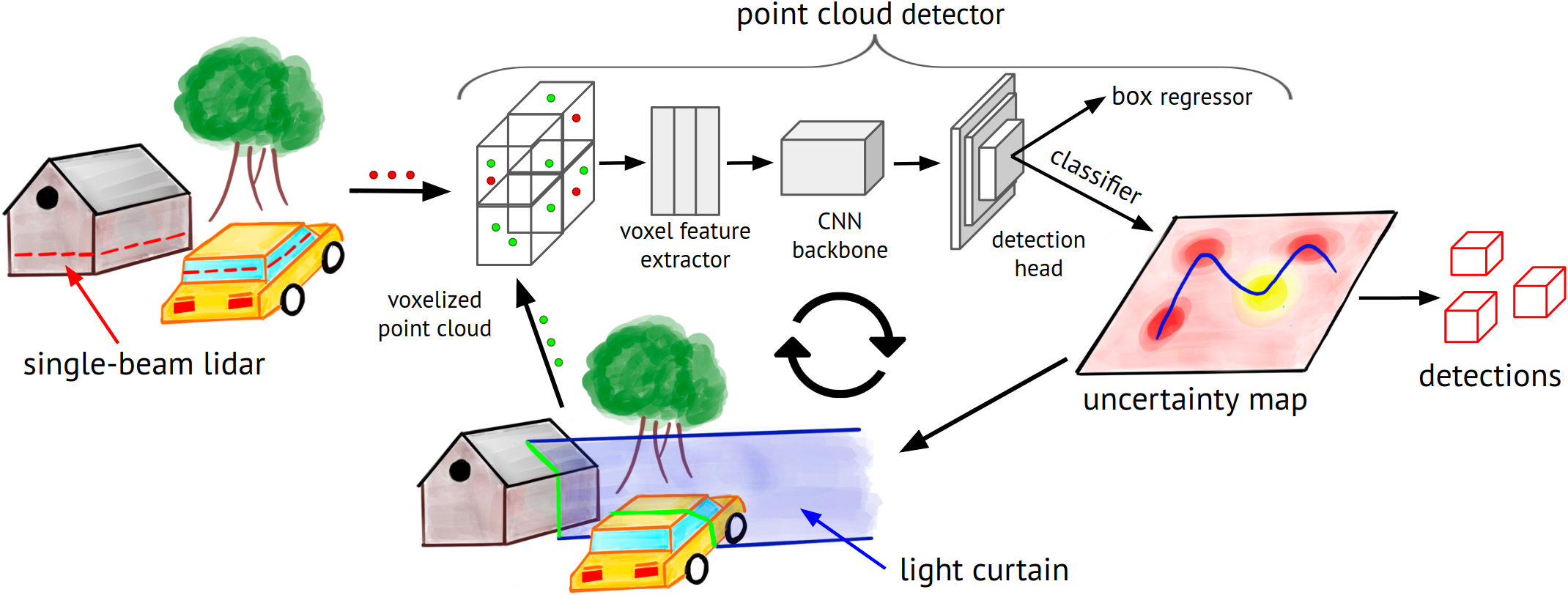}
    \caption{\textit{Our method for detecting objects using light curtains.} An inexpensive single-beam lidar input is used by a 3D detection network to obtain rough initial estimates of object locations. The uncertainty of the detector is used to optimally place a light curtain that covers the most uncertain regions. The points detected by the light curtain (shown in green in the bottom figure) are input back into the detector so that it can update its predictions as well as uncertainty. The new uncertainty maps can again be used to place successive light curtains in an iterative manner, closing the loop.}
\label{fig:pipeline}
\vspace{-10pt}
\end{figure}

Our aim is to use light curtains for detecting objects in a 3D scene. The overall approach is illustrated in Fig.~\ref{fig:pipeline}. We use a voxel-based point cloud detector~\cite{yan2018second} and train it to use light curtain data without any architectural changes. The pipeline illustrated in Fig.~\ref{fig:pipeline} proceeds as follows.

To obtain an initial set of object detections, we use data from an inexpensive single-beam LiDAR as input to the detector. This produces rough estimates of object locations in the scene. Single-beam LiDAR is inexpensive because it consists of only one laser beam as opposed to 64 or 128 beams that are common in autonomous driving. The downside is that the data from the single beam contains very few points; this results in inaccurate detections and high uncertainty about object locations in the scene (see Fig.~\ref{fig:pull}b).


Alongside bounding box detections, we can also extract from the detector an ``uncertainty map" (explained in  Sec.~\ref{sec:uncertainty-extraction}). We then use light curtains, placed in regions guided by the detector's uncertainty, to collect more data and iteratively refine the object detections. In order to get more data from the regions the detector is most uncertain about, we derive an information-gain based objective function that sums the uncertainties along the light curtain control points (Sec.~\ref{sec:objective} and Appendix~\ref{app:info-gain}),
and we develop a constrained optimization algorithm that places the light curtain to maximize this objective (Sec.~\ref{sec:dp}).


Once the light curtain is placed, it returns a dense set of points where the curtain intersects with visible objects in the scene. We maintain a \textit{unified point cloud}, which we define as the union of all points observed so far. The unified point cloud is initialized with the points from the single-beam LiDAR. Points from the light curtain are added to the unified point cloud and this data is input back into the detector. Note that the input representation for the detector remains the same (point clouds), enabling the use of existing state-of-the-art point cloud detection methods without any architectural modifications. 

As new data from the light curtains are added to the unified point cloud and input to the detector, the detector refines its predictions and improves its accuracy. Furthermore, the additional inputs cause the network to update its uncertainty map; the network may no longer be uncertain about the areas that were sensed by the light curtain. Our algorithm uses the new uncertainty map to generate a new light curtain placement. We can iteratively place light curtains to cover the current uncertain regions and input the sensed points back into the network, closing the loop and iteratively improving detection performance.



\subsection{Extracting uncertainty from the detector}
\label{sec:uncertainty-extraction}
The standard pipeline for 3D object detection~\cite{zhou2018voxelnet,yan2018second,lang2019pointpillars} proceeds as follows. First, the ground plane (parallel to the $xz$-plane)
is uniformly tiled with ``anchor boxes"; these are reference boxes used by a 3D detector to produce detections. They are located on points in a uniformly discretized grid $G = [x_\text{min}, x_\text{max}]\times[z_\text{min}, z_\text{max}]$. For example, a $[-40\text{m}, 40\text{m}] \times [0\text{m}, 70.4\text{m}]$ grid is used for detecting cars in KITTI~\cite{geiger2013vision}. A 3D detector, which is usually a binary detector, takes a point cloud as input, and produces a binary classification score $p \in [0, 1]$ and bounding box regression offsets for every anchor box. The score $p$ is the estimated probability that the anchor box contains an object of a specific class (such as car/pedestrian). The detector produces a detection for that anchor box if $p$ exceeds a certain threshold. If so, the detector combines the fixed dimensions of the anchor box with its predicted regression offsets to output a detection box.

We can convert the confidence score to binary entropy $H(p) \in [0, 1]$ where $H(p) = -p\log_2 p - (1-p)\log_2(1-p)$. Entropy is a measure of the detector's uncertainty about the presence of an object at the anchor location. Since we have an uncertainty score at uniformly spaced anchor locations parallel to the $xz$-plane, they form an ``uncertainty map'' in the top-down view. We use this uncertainty map to place light curtains.

\subsection{Information gain objective}
\label{sec:objective}
\newcommand{\A}{\mathbf{A}}

Based on the uncertainty estimates given by Sec.~\ref{sec:uncertainty-extraction}, our method determines how to place the light curtain to sense the most uncertain/ambiguous regions.
It seems intuitive that sensing the locations of highest detector uncertainty can provide the largest amount of information from a single light curtain placement, towards improving detector accuracy.
As discussed in Sec.~\ref{sec:background}, a single light curtain placement is defined by a set of $T$ control points $\{\X_t\}_{t=1}^T$. The light curtain will be placed to lie vertically on top of these control points.
To define an optimization objective, we use the framework of information gain (commonly used in next-best view methods; see Sec.~\ref{sec:related-work-nbv}) along with some simplifying assumptions (see Appendix~\ref{app:info-gain}). We show that under these assumptions, placing a light curtain to maximize information gain (a mathematically defined information-theoretic quantity) is equivalent to maximizing the objective $J(\X_1, \dots, \X_T) = \sum_{t=1}^T H(\X_t)$, where $H(\X)$ is the binary entropy of the detector's confidence at the anchor location of $\X$. When the control point $\X$ does not exactly correspond to an anchor location, we impute $H(\X)$ by nearest-neighbor interpolation from the uncertainty map. Please see Appendix~\ref{app:info-gain} for a detailed derivation.

\subsection{Optimal light curtain placement}
\label{sec:dp}

In this section, we will describe an exact optimization algorithm to maximize the objective function $J(\X_1, \dots, \X_T) = \sum_{t=1}^T H(\X_t)$.\\
\textbf{Constrained optimization}: The control points $\{\X_t\}_{t=1}^T$, where each $\X_t$ lies on the the camera ray $\R_t$, must be chosen to satisfy the physical constraints of the light curtain device: $|\theta(\X_{t+1}) - \theta(\X_t)| \leq \Delta \theta_\text{max}$ (see Sec.~\ref{sec:background}: light curtain constraints). Hence, this is a constrained optimization problem. We discretize the problem by considering a dense set of $N$ discrete, equally spaced points $\D_t = \{\X_t^{(n)}\}_{n=1}^N$ on each ray $\R_t$. We will assume that $\X_t \in \D_t$ for all $1 \leq t \leq T$ henceforth unless stated otherwise. We use $N=80$ in all our experiments which we found to be sufficiently large. Overall, the optimization problem can be formulated as:
\begin{align}
    &\arg \max_{\{\X_t\}_{t=1}^T} \sum_{t=1}^T H(\X_t)\\
    &\ \ \ \ \ \ \ \text{where}\ \X_t \in \D_t\ \forall 1 \leq t \leq T\\
    &\ \ \ \ \ \ \ \text{subject to}\ |\theta(\X_{t+1}) - \theta(\X_t)| \leq \dtmax,\ \forall 1 \leq t < T
    \label{eq:constraint}
\end{align}
\begin{figure}[t]
    \centering
    \subfloat[]{
        \includegraphics[trim=0 3 1 3,clip,width=0.45\textwidth]{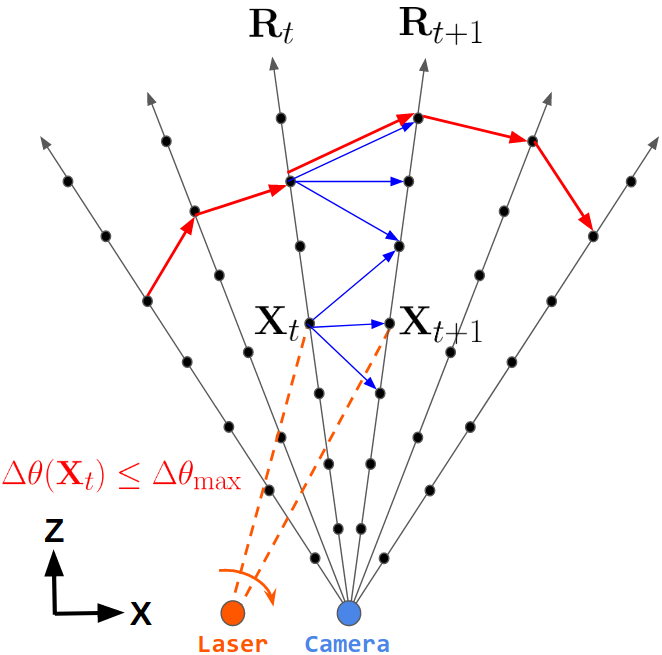}
    }
    \subfloat[]{
        \includegraphics[trim=11 9 15 0,clip,width=0.47\textwidth]{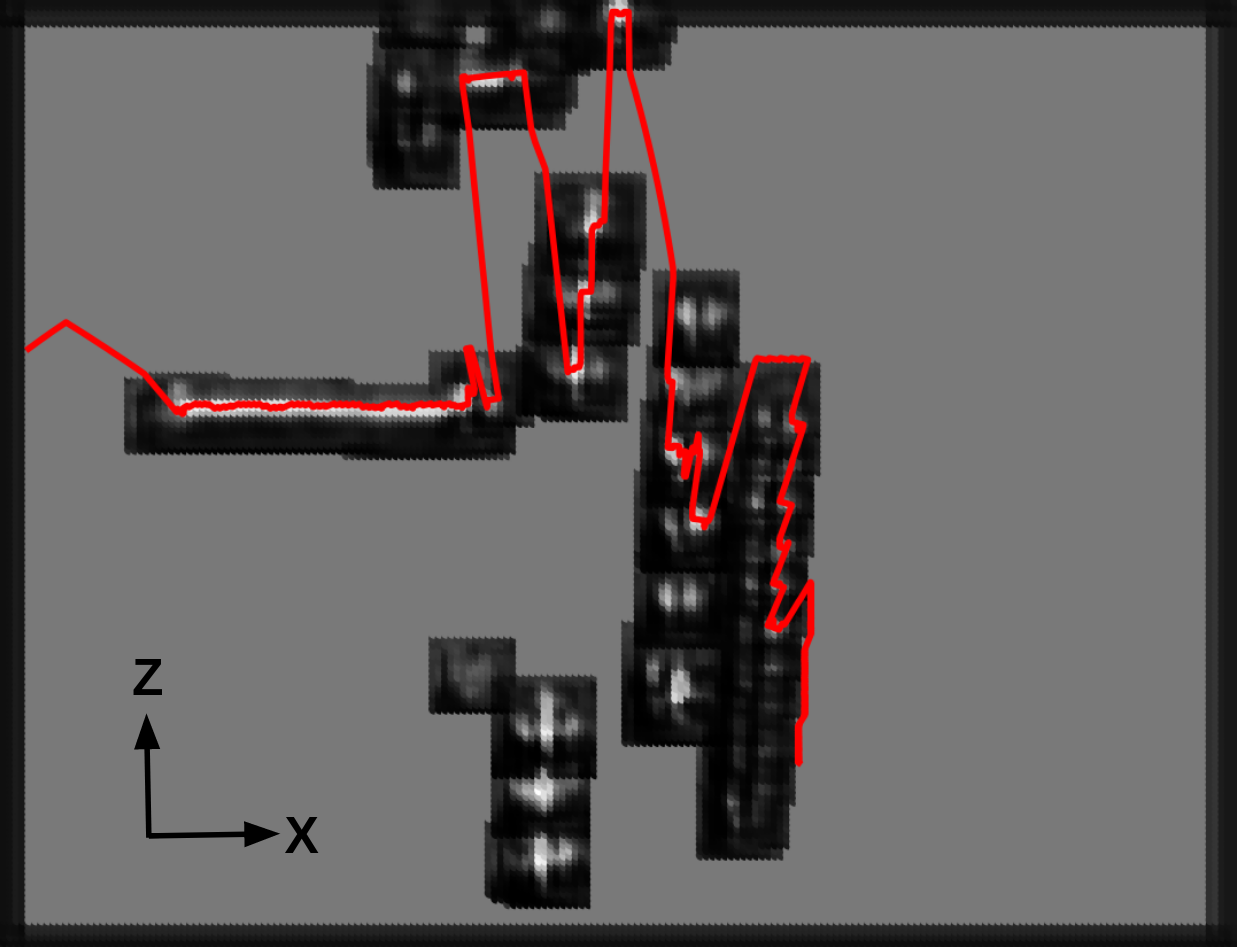}
    }
    \caption{(a) Light curtain constraint graph. Black dots are nodes and blue arrows are the edges of the graph. The optimized light curtain profile is depicted as red arrows. (b) Example uncertainty map from the detector and optimized light curtain profile in red. Black is lowest uncertainty and white is highest uncertainty. The optimized light curtain covers the most uncertain regions.}
    \label{fig:dp}
    \vspace{-10pt}
\end{figure}

\noindent \textbf{Light Curtain Constraint Graph:} we encode the light curtain constraints into a graph, as illustrated in Figure~\ref{fig:dp}.
Each black ray corresponds to a camera ray. Each black dot on the ray is a vertex in the constraint graph. It represents a candidate control point and is associated with an uncertainty score. Exactly one control point must be chosen per camera ray. The optimization objective is to choose such points to maximize the total sum of uncertainties. An edge between two control points indicates that the light curtain is able to transition from one control point $\X_t$ to the next, $\X_{t+1}$ without violating the maximum velocity light curtain constraints. 
Thus, the maximum velocity constraint (Eqn.~\ref{eq:constraint}) can be specified by restricting the set of edges (depicted using blue arrows). We note that the graph only needs to be constructed once and can be done offline.\\
\textbf{Dynamic programming for constrained optimization:} The number of possible light curtain placements, $|\D_1 \times \dots \times \D_T| = N^T$, is exponentially large, which prevents us from searching for the optimal solution by brute force. However, we observe that the problem can be decomposed into simpler subproblems. In particular, let us define $J^*_t(\X_t)$ as the optimal sum of uncertainties  of the \textit{tail subproblem} starting from $\X_t$ i.e.
\begin{align}
    J_t^*(\X_t) = &\max_{\X_{t+1}, \dots, \X_T} H(\X_t) + \sum_{k=t+1}^T H(\X_k); \\
    &\text{subject to}\ |\theta(\X_{k+1}) - \theta(\X_k)| \leq \dtmax,\ \forall\ t \leq k < T
\end{align}
If we were able to compute $J^*_t(\X_t)$, then this would help in solving a more complex subproblem using recursion:
we observe that $J^*_t(\X_t)$ has the property of \emph{optimal substructure,} i.e. the optimal solution of $J_{t-1}^*(\X_{t-1})$ can be computed from the optimal solution of $J_t^*(\X_t)$ via
\begin{equation}
\begin{aligned}
    J_{t-1}^*(\X_{t-1}) = H(\X_{t-1}) + &\max_{\X_t \in \D_t} J_t^*(\X_{t})\\
    &\text{subject to}\ |\theta(\X_t) - \theta(\X_{t-1})| \leq \dtmax
\label{eq:bellman}
\end{aligned}
\end{equation}
Because of this optimal substructure property, we can solve for $J_{t-1}^*(\X_{t-1})$ via dynamic programming.  We also note that the solution to $\max_{\X_1} J_1^*(\X_1)$ is the solution to our original constrained optimization problem (Eqn. 1-3).

We thus perform the dynamic programming optimization as follows:
the recursion from Eqn.~\ref{eq:bellman} can be implemented by first performing a backwards pass, starting from $T$ and computing $J_t^*(\X_t)$ for each $\X_t$. Computing each $J_t^*(\X_t)$ takes only $O(B_\text{avg})$ time where $B_\text{avg}$ is the average degree of a vertex (number of edges starting from a vertex) in the constraint graph, since we iterate once over all edges of $\X_t$ in Eqn.~\ref{eq:bellman}. Then, we do a forward pass, starting with $\arg \max_{\X_1 \in \D_1} J_1^*(\X_1)$ and for a given $\X^*_{t-1}$, choosing $\X^*_t$ according to Eqn.~\ref{eq:bellman}. Since there are $N$ vertices per ray and $T$ rays in the graph, the overall algorithm takes $O(NTB_\text{avg})$ time; this is a significant reduction from the $O(N^T)$ brute-force solution.\\
\textbf{Hierarchical optimization objective for smoothness:} If two light curtain placements produce the same sum of uncertainties, which one should we prefer? We propose a hierarchical optimization objective that prefers smoother light curtains. We show that this also has the optimal substructure property and can be optimized in a very similar manner (see Appendix~\ref{app:hierarchical-opt} for details).

\subsection{Training active detector with online training data generation}
\label{sec:online-training}


We now describe our approach to train 3D point cloud detectors with data from light curtains and single-beam lidar. At each training iteration $t$, we retrieve a scene $S_t$ from the training dataset. To create the input point cloud, we choose to either use the single-beam LiDAR data or $k$ light curtain placements ($1 \leq k \leq K$), each of them with equal probability. For generating the $k$-th light curtain data, we start with the single-beam LiDAR point cloud. Then we successively perform a forward pass through the detector network with the current weights to obtain an uncertainty map. We compute the optimal light curtain placement for this map, gather points returned from placing this curtain, and finally, fuse the points back into the input point cloud. This cycle is repeated $k$ times to obtain the input point cloud to train on. Generating light curtain data in such an \emph{online} fashion ensures that the input distribution doesn't diverge from the network weights during the course of training. See Appendix~\ref{app:training-algo} for more algorithmic details and an ablation experiment that evaluates the importance of online training data generation. 

\section{Experiments}
\label{sec:experiments}

\textbf{Datasets:}
To evaluate our algorithm, we need dense ground truth depth maps to simulate an arbitrary placement of a light curtain. However, standard autonomous driving datasets, such as  KITTI~\cite{geiger2013vision} and nuScenes~\cite{caesar2019nuscenes},  contain only sparse LiDAR data, and hence the data is not suitable to accurately simulate a dense light curtain to evaluate our method. To circumvent this problem, we demonstrate our method on two synthetic datasets that provide dense ground truth depth maps, namely the Virtual KITTI~\cite{vkitti} and SYNTHIA~\cite{synthia} datasets.  
Virtual KITTI is a photo-realistic synthetic video dataset designed for video understanding tasks~\cite{vkitti}. It contains 21,160 frames (10,630 unique depth maps) generated from five different virtual worlds in urban driving settings design to closely resemble five scenes in the KITTI dataset, under different camera poses and weather conditions. It provides ground truth depth maps and 3D bounding boxes. We use four scenes (ids: 0001, 0006, 0018, 0020) as our training set, and one scene (id: 0002) as our test set.

We also use the latest version of the SYNTHIA dataset~\cite{synthia} designed for active learning purposes. It is a large dataset containing photo-realistic scenes from urban driving scenarios, and provides ground truth depth and 3D bounding box annotations. It contains 191 training scenes (\mytilde 96K frames) and 97 test scenes (\mytilde 45K frames).

\textbf{Evaluation metrics:} 
We evaluate using common 3D detection metrics: mean average precision (mAP) of 3D bounding boxes (denoted as 3D mAP) and of 2D boxes in the bird's eye view (denoted as BEV mAP). We also evaluate using two different IoU overlap thresholds of 0.5 and 0.7 between detection boxes and ground-truth boxes to be considered true positives.

Our experiments demonstrate the following:
First, we show that our method for successive placement of light curtains improves detection performance; particularly, there is a significant increase between the performance of single-beam LiDAR and the performance after placing the first light curtain. We also  compare our method to multiple ablations and alternative placement strategies that demonstrate that each component of our method is crucial to achieve good performance. Finally, we show that our method can generalize to many more light curtain placements at test time than the method was trained on.
\addition{In the appendix, we perform further experiments that include evaluating the generalization of our method to noise in the light curtain data, an ablation experiment for training with online data generation (Sec.~\ref{sec:online-training}), and efficiency analysis.}

\subsection{Comparison with varying number of light curtains}

We train our method using online training data generation simultaneously on data from single-beam LiDAR and one, two, and three light curtain placements. We perform this experiment for both the Virtual KITTI and SYNTHIA datasets. The accuracies on their tests sets are reported in Table~\ref{table:main-results}.

\begin{table*}[h!]
  \centering
  \begin{tabular}{?c?c|c|c|c?c|c|c|c?} 
   \Xhline{0.8pt}
   & \multicolumn{4}{c?}{\bf{Virtual KITTI}} & \multicolumn{4}{c?}{\bf{SYNTHIA}}\\
   \hline
   & \multicolumn{2}{c|}{3D mAP} & \multicolumn{2}{c?}{BEV mAP} & \multicolumn{2}{c|}{3D mAP} & \multicolumn{2}{c?}{BEV mAP}\\
   \hline
   & 0.5 IoU & 0.7 IoU & 0.5 IoU & 0.7 IoU & 0.5 IoU & 0.7 IoU & 0.5 IoU & 0.7 IoU\\
   \Xhline{0.8pt}
   Single Beam Lidar & 39.91 & 15.49 & 40.77 & 36.54 & 60.49 &  47.73 &   60.69 &   51.22\\
   \hline
   \makecell{Single Beam Lidar\\(separate model)} & 42.35 & 23.66 & 47.77 & 40.15 & 60.69 &  48.23 &   60.84 &   57.98\\
   \hline
   1 Light Curtain & 58.01 & 35.29 & 58.51 & 47.05 & 68.79 &  55.99 &   68.97 &   59.63\\
   \hline
   2 Light Curtains & 60.86 & 37.91 & 61.10 & 49.84 & 69.02 &  57.08 &   69.17 &   67.14\\
   \hline
   3 Light Curtains & \bf{68.52} & \bf{38.47} & \bf{68.82} & \bf{50.53} & \bf{69.16} &  \bf{57.30} &   \bf{69.25} &   \bf{67.25}\\
   \Xhline{0.8pt}
  \end{tabular}
  \caption{Performance of the detector trained with single-beam LiDAR and up to three light curtains. Performance improves with more  light curtain placements, with a significant jump at the first light curtain placement.}
  \label{table:main-results}
  \vspace{-15pt}
\end{table*}

Note that there is a significant and consistent increase in the accuracy between single-beam LiDAR performance and the first light curtain placement (row 1 and row 3). This shows that actively placing light curtains on the most uncertain regions can improve performance over a single-beam LiDAR \addition{that performs fixed scans}. Furthermore, placing more light curtains consistently improves detection accuracy.

As an ablation experiment, we train a separate model only on single-beam LiDAR data (row 2), for the same number of training iterations. This is different from row 1 which was trained with both single beam LiDAR and light curtain data but evaluated using only data for a single beam LiDAR.  Although training a model with only single-beam LiDAR data (row 2) improves performance over row 1, it is still significantly outperformed by our method which uses data from light curtain placements.

\textbf{Noise simulations}: In order to simulate noise in the real-world sensor, we perform experiments with added noise in the light curtain input. We demonstrate that the results are comparable to the noiseless case, indicating that our method is robust to noise and is likely to transfer well to the real world. Please see Appendix~\ref{app:noise} for more details.

\subsection{Comparison with alternative light curtain placement strategies}

\begin{table*}[!b]
  \centering
  \begin{tabular}{?c?c|c|c|c?c|c|c|c?} 
   \Xhline{0.8pt}
   & \multicolumn{4}{c?}{\bf{Virtual KITTI}} & \multicolumn{4}{c?}{\bf{SYNTHIA}}\\
   \hline
   & \multicolumn{2}{c|}{3D mAP} & \multicolumn{2}{c?}{BEV mAP} & \multicolumn{2}{c|}{3D mAP} & \multicolumn{2}{c?}{BEV mAP}\\
   \hline
   & .5 IoU & .7 IoU & .5 IoU & .7 IoU & .5 IoU & .7 IoU & .5 IoU & .7 IoU\\
   \Xhline{0.8pt}
   Random & 41.29 &  17.49 &   46.65 &   38.09 & 60.43 &  47.09 &   60.66 & 58.14 \\
   \hline
   Fixed depth - 15m& 44.99 &  22.20 &   46.07 &   38.05 & 60.74 &  48.16 & 60.89 & 58.48 \\
   \hline
   Fixed depth - 30m& 39.72 &  19.05 &   45.21 &   35.83 & 60.02 &  47.88 & 60.23 & 57.89 \\
   \hline
   Fixed depth - 45m& 39.86 &  20.02 &   40.61 &   36.87 & 60.23 & 48.12 & 60.43 & 57.77 \\
   \hline
   \makecell{Greedy Optimization\\(Randomly break ties)}& 37.40 &  19.93 &   42.80 &   35.33 & 60.62 & 47.46 & 60.83 & 58.22\\
   \hline
   \makecell{Greedy Optimization\\(Min laser angle change)}& 39.20 &  20.19 &   44.80 &   36.94 & 60.61 & 47.05 & 60.76 & 58.07 \\
   \hline
   \makecell{Frontoparallel +\\Uncertainty} & 39.41 &  21.25 &   45.10 &   37.80 & 60.36 &  47.20 &   60.52 &   58.00 \\
   \hline
   \textbf{Ours} & \bf{58.01} &  \bf{35.29} &   \bf{58.51} &   \bf{47.05} & \bf{68.79} &  \bf{55.99} &   \bf{68.97} &   \bf{59.63} \\
   \Xhline{0.8pt}
  \end{tabular}
  \caption{Baselines for alternate light curtain placement strategies, trained and tested on (a) Virtual KITTI and (b) SYNTHIA datasets. Our dynamic programming optimization approach significantly outperforms all other strategies.}
  \label{table:baselines}
\end{table*}

In our approach, light curtains are placed by maximizing the coverage of uncertain regions using a dynamic programming optimization. How does this compare to other strategies for light curtain placement? We experiment with several baselines:
\begin{enumerate}
  \item \textit{Random}: we place frontoparallel light curtains at a random $z$-distance from the sensor, ignoring the detector's uncertainty map.
  \item \textit{Fixed depth}: we place a frontoparallel light curtain at a fixed $z$-distance (15m, 30m, 45m) from the sensor, ignoring the detector's uncertainty map.
  \item \textit{Greedy optimization}: this baseline tries to evaluate the benefits of using a dynamic programming optimization. Here, we use the same light curtain constraints described in Section~\ref{sec:dp} (Figure~\ref{fig:dp}(a)). We greedily select the next control point based on local uncertainty instead of optimizing for the future sum of uncertainties. Ties are broken by (a) choosing smaller laser angle changes, and (b) randomly.
  \item \textit{Frontoparallel + Uncertainty}: Our optimization process finds light curtains with flexible shapes. What if the shapes were constrained to make the optimization problem easier? If we restrict ourselves to frontoparallel curtains, we can place them at the $z$-distance of maximum uncertainty by simply summing the uncertainties for every fixed value of $z$.
\end{enumerate}

\begin{figure}[t]
  \centering
  \subfloat[Generalization in Virtual KITTI]{
      \includegraphics[trim=12 13 10 10,clip,width=0.485\textwidth]{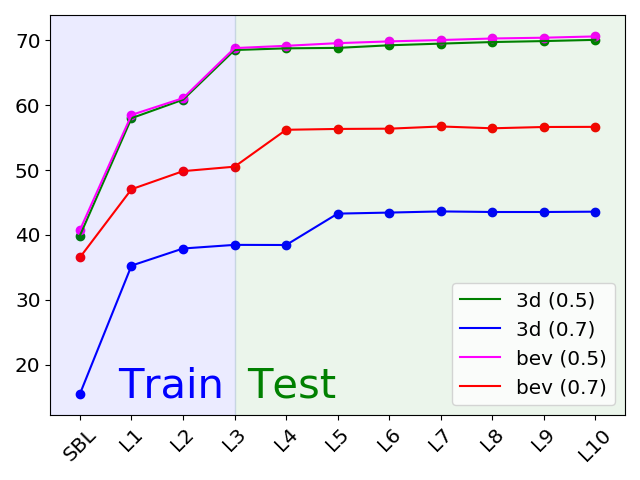}
  }
  \subfloat[Generalization in SYNTHIA]{
    \includegraphics[trim=12 13 10 10,clip,width=0.485\textwidth]{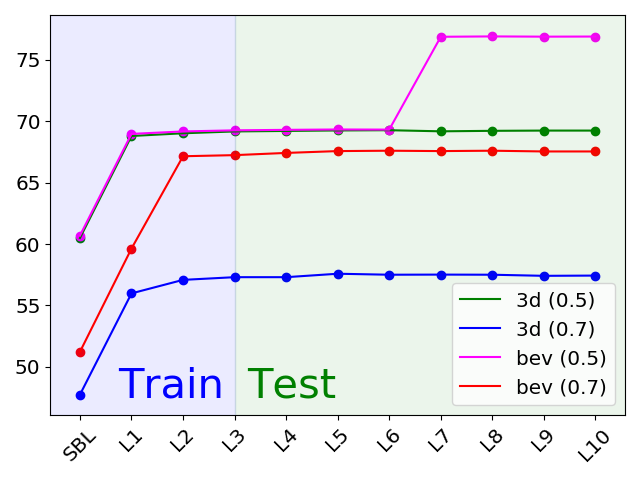}
  }
  \caption{\textit{Generalization to many more light curtains than what the detector was trained for}. We train using online data generation on single-beam lidar and only 3 light curtains. We then test with placing 10 curtains, on (a) Virtual KITTI, and (b) SYNTHIA. Performance continues to increase monotonically according to \addition{multiple} metrics. Takeaway: one can safely place more light curtains at test time and expect to see sustained improvement in accuracy.}
  \label{fig:lcg}
  \vspace{-10pt}
\end{figure}

The results on the Virtual KITTI and SYNTHIA datasets are shown in Table~\ref{table:baselines}. Our method significantly and consistently outperforms all baselines. This empirically demonstrates the value of using dynamic programming for light curtain placement to improve object detection performance.

\begin{figure}
  \centering
  \includegraphics[trim=3 0 0 0,clip,width=\textwidth]{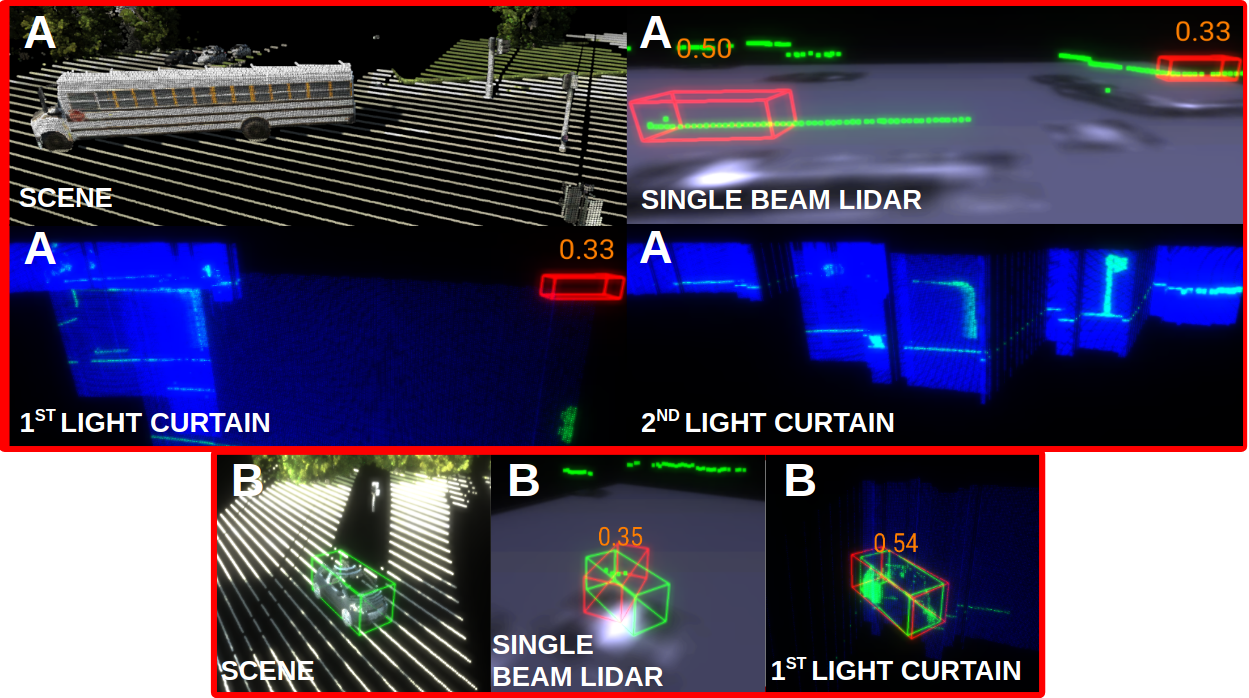}
  \caption{\textit{Successful cases:} Other type of successful cases than Fig.~\ref{fig:pull}. In (A), \addition{the} single-beam LiDAR incorrectly detects a bus and a piece of lawn as false positives. They get eliminated successively after placing \addition{the} first and second light curtains. In (B), \addition{the} first light curtain fixes misalignment in the bounding box predicted by the single beam LiDAR.} 
  \label{fig:qual-success}
\end{figure}

\subsection{Generalization to successive light curtain placements}

If we train a detector using our online light curtain data generation approach for $k$ light curtains, can the performance generalize to more than $k$ light curtains? Specifically, if we continue to place light curtains beyond the number trained for, will the accuracy continue improving? We test this hypothesis by evaluating on 10 light curtains, many more than the model was trained for (3 light curtains). Figure~\ref{fig:lcg} shows the performance as a function of the number of light curtains. We find that in both Virtual KITTI and SYNTHIA, the accuracy monotonically improves with the number of curtains.

This result implies that a priori one need not worry about how many light curtains will be placed at test time. If we train on only 3 light curtains, we can place many more light curtains at test time; our results indicate that the performance will keep improving.

\subsection{Qualitative analysis}
\begin{figure}
  \centering
  \includegraphics[trim=0 0 2 0,clip,width=\textwidth]{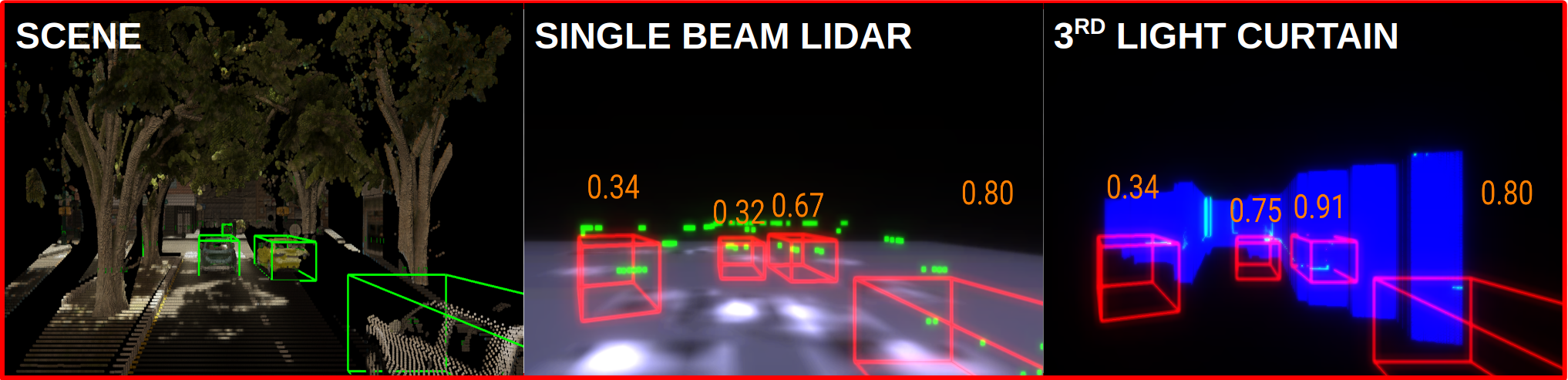}
  \caption{\textit{Failure cases:} The predominant failure mode is that the single beam LiDAR detects a false positive which is not removed by light curtains because the detector is overly confident in its prediction (so the estimated uncertainty is low). \textit{Middle}: Falsely detecting a tree to be a car. \textit{Right}: After three light curtains, the detection persists because light curtains do not get placed on this false positive.}
  \label{fig:qual-failure}
\end{figure}

We visualized a successful case of our method in Fig.~\ref{fig:pull}. This is an example where our method detects false negatives missed by the single-beam LiDAR. We also show two other types of successful cases  where light curtains remove false positive detections and fix misalignment errors in Figure~\ref{fig:qual-success}. In Figure~\ref{fig:qual-failure}, we show the predominant failure case of our method. See captions for more details.

The predominant failure case of our method is when the LiDAR makes a mistake, such as a false positive in Fig.~\ref{fig:qual-failure}, but the light curtain fails to be placed in that region to fix the mistake. This happens when the detector makes a mistake but is very confident in its prediction; in such a case, the estimated uncertainty for this prediction will be low and a light curtain may not be placed at this location. In this particular example shown, after six light curtain placements, a light curtain eventually gets placed at the location of the false positive and the detector fixes its mistake. However, in other examples, a light curtain might never be placed at the location of the incorrect detection, due to an overly confident (but incorrect) prediction.
\section{Conclusions}

In this work, we develop a method to use light curtains, an actively controllable resource-efficient sensor, for object recognition in static scenes. We propose to use a 3D object detector's prediction uncertainty as a guide for deciding where to sense. By encoding the constraints of the light curtain into a graph, we show how to optimally and feasibly place a light curtain that maximizes the coverage of uncertain regions. We are able to train an active detector that interacts with light curtains to iteratively and efficiently sense parts of scene in an uncertainty-guided manner, successively improving detection accuracy. \addition{We hope this work pushes towards replacing expensive multi-beam LiDAR systems with inexpensive controllable sensors, enabled by} designing perception algorithms for autonomous driving that integrate sensing and recognition.




\section*{Acknowledgements}

We thank Matthew O'Toole for feedback on the initial draft of this paper. This material is based upon work supported by the National Science Foundation under Grants No.  IIS-1849154, IIS-1900821 and by the United States Air Force and DARPA under Contract No. FA8750-18-C-0092.

%
%
\bibliographystyle{splncs04}
\bibliography{main}

\appendix
\title{
  {\huge Appendix}
  \\\ \\
  Active Perception using Light Curtains for Autonomous Driving
}

\author{Siddharth Ancha \and
Yaadhav Raaj \and
Peiyun Hu \and\\
Srinivasa G. Narasimhan \and
David Held}
\authorrunning{S. Ancha et al.}
\titlerunning{Active Perception using Light Curtains for Autonomous Driving}
%
\institute{Carnegie Mellon University, Pittsburgh PA 15213, USA\\
\email{\{sancha,ryaadhav,peiyunh,srinivas,dheld\}@andrew.cmu.edu}}

\maketitle

\section{Information gain objective}
\label{app:info-gain}

In this section, we derive the optimization objective used in Sections~\ref{sec:objective} and~\ref{sec:dp}, from a perspective of maximizing information gain. Information gain is a well-defined mathematical quantity, and choosing \addition{sensing} actions to maximize information gain has been used as the basis of many works on next-best view planning (see Sec.~\ref{sec:background}).

We will first describe some notation, and make two simplifying assumptions in order to derive our objective as an approximation of information gain.

\subsection{Notation}
\begin{itemize}
    \item[\textbullet] The detector predicts the probability of a detection at every anchor box location. Let there be a total of $K$ discrete anchor box location, which are usually organized as a regular 2D-grid (see Sec.~\ref{sec:uncertainty-extraction}). Let $\A_k$ denote the $k$-th anchor box, where $1 \leq k \leq K$. Define $\A=\{\A_k\}_{k=1}^K$ to be the vector of all anchor boxes.
    \item[\textbullet] Let $D_{\A_k}$ be a binary random variable denoting whether a detection exists at $\A_k$. $D_{\A_k} \in \{0, 1\}$; it is $0$ if there is no detection at $\A_k$, and $1$ if there is. Define $D_\A=\{D_{\A_k}\}_{k=1}^K$.
    \item[\textbullet] Given a unified point cloud $C$, an inference algorithm (in this case, the detector) outputs a probability distribution $P(D_\A\ |\ C)$ over all possible detection states $D_\A \in \{0, 1\}^K$. Denote by $P(D_{\A_k})$ the marginal probability distribution of detection at $\A_k$.
    \item[\textbullet] As discussed in Sec.~\ref{sec:background}, a single light curtain placement is defined by a set of control points $L=\{\X_t\}_{t=1}^T$. The light curtain will be placed to lie vertically on top of these control points. The 3D points sensed by this light curtain are fused back into $C$, to obtain an updated unified point cloud $C'$. We assume for now that the control points $\X_t$ correspond to some anchor box locations.

\end{itemize}

\subsection{Assumptions}
We now make the following assumptions:

\begin{enumerate}
    \item \textit{Detections probabilities across locations are independent}.\\
    That is, $P(D_\A\ |\ C) = \prod_{k=1}^K P(D_{\A_k}\ |\ C)$. This is a reasonable assumption, since the probability of detections at one location should be unaffected by detections in other locations. A consequence of this assumption is that the overall entropy $H(D_\A\ |\ C)$ can be written as the sum of entropies over individual anchor locations i.e. $H(D_\A\ |\ C) = \sum_{k=1}^K H(D_{\A_k}\ |\ C)$ (since the entropy of independent random variables is the sum of their individual entropies).
    \item \textit{Light curtain sensing resolves uncertainty fully but locally}.\\
    After placing $L = \{\X_t\}_{t=1}^T$, updating the unified point cloud to $C'$, re-running the detector, and obtaining a new probability distribution of the updated detections $D'_\A$, the following hold.
    \begin{enumerate}
        \item The uncertainty of locations covered by the curtain reduces to zero:\\
        $P(D'_{\A_k}\ |\ C') \in \{0, 1\}$ for all $\A_k \in L$.
        \item The uncertainty of all the other locations remains unchanged:\\
        $P(D'_{\A_k}\ |\ C') = P(D_{\A_k}\ |\ C)$ for all $\A_k \not\in L$.
    \end{enumerate}
\end{enumerate}
Assumptions 1 and 2 imply that the entropy of the updated distribution is given by (here $K$ is the total number of anchor locations, and $T$ is the number of locations that the light curtain lies on).
\begin{align*}
    H(D'_\A\ |\ C') &= \sum_{k=1}^K H(D'_{\A_k}\ |\ C')\\
    &= \sum_{\A_k \in L} \underbrace{H(D'_{\A_k}\ |\ C')}_{=\ 0 \text{ as } P(D'_{\A_k}|C') \in \{0, 1\}} + \sum_{\A_k \not\in L} \underbrace{H(D'_{\A_k}\ |\ C')}_{=\ H(D_{\A_k}\ |\ C)}\\
    &= \sum_{\A_k} H(D_{\A_k}\ |\ C) - \sum_{\A_k \in L} H(D_{\A_k}\ |\ C)\\
    &= \sum_{k=1}^K H(D_{\A_k}\ |\ C) - \sum_{\A_k \in L} H(D_{\A_k}\ |\ C)\\
    &=H(D_\A\ |\ C) - \sum_{t=1}^T H(D_{\X_t}\ |\ C)    
\end{align*}
The information gain, which is essentially a difference between the prior and updated entropies, is
\begin{align*}
\text{Information Gain} &= H(D_{\A}\ |\ C) - H(D'_{\A}\ |\ C')\\
&= H(D_{\A}\ |\ C) - \Big( H(D_{\A}\ |\ C) - \sum_{t=1}^T H(D_{\X_t}\ |\ C) \Big)\\
&= \sum_{t=1}^T H(D_{\X_t}\ |\ C)
\end{align*}

\textbf{Optimization objective}: This leads us to an optimization objective where maximizing information gain is equivalent to simply maximizing the sum of uncertainties (binary entropies) over the control points the curtain lies on.
The maximization objective then becomes: $J(\X_1, \dots, \X_T) = \sum_{t=1}^T H(\X_t)$, where $H(\X)$ is the binary entropy of the detector’s confidence at the location of $\X$.
\section{Hierarchical optimization objective for smoothness}
\label{app:hierarchical-opt}

Section~\ref{sec:dp} described an efficient algorithm for optimally placing light curtains to maximize coverage of high uncertainy regions. However, if two valid light curtain placements $\{\X'_t\}_{t=1}^T, \{\X''_t\}_{t=1}^T$ have equal sum of uncertainties, which one should we prefer? Distinct light curtain placements can have equal sums of uncertainties due to regions where the detector uncertainty is uniform. In such cases, we can choose to prefer curtains that are \textit{smooth}, i.e. the laser angle has to change the least on average. We define a hierarchical objective function that ranks two placements as follows:
\[ J_H(\{\X'_t\}_{t=1}^T) \geq J_H(\{\X''_t\}_{t=1}^T) \text{ iff }
    \begin{cases}
        J(\{\X'_t\}_{t=1}^T) > J(\{\X''_t\}_{t=1}^T)\\
        \text{or}\\
        \begin{cases}
        J(\{\X'_t\}_{t=1}^T) = J(\{\X''_t\}_{t=1}^T)\\
        \text{\ \ \ \ \ \ \ \ \ \ \ \ \ \ \ and }\\
        \sum_{t=1}^{T-1} |\theta(\X'_{t+1}) - \theta(\X'_{t})|^2 \\\leq \sum_{t=1}^{T-1} |\theta(\X''_{t+1}) - \theta(\X''_{t})|^2
        \end{cases}
    \end{cases}
  \]
This hierarchical objective prefers light curtains that cover a higher sum of uncertainties. But if two curtains have the same sum, this objective prefers the one with a lower sum of squared laser angle deviations. We note that this hierarchical objective $J_H(\X_1, \dots, \X_T)$ also satisfies optimal substructure. In fact, it obeys the same recursive optimality equation as Equation~\ref{eq:bellman}. Hence, it can be accommodated by our approach with minimal modification to our algorithm. Additionally, it can be executed with no additional overhead in $O(NTB_\text{avg})$ time, and leads to smoother light curtains.
\section{Training active detection with online light curtain data generation}
\label{app:training-algo}

In this section, we expand on the details of our method to train the detector described in Section~\ref{sec:online-training}. Note that we use the same detector to process data from the single beam LiDAR and all subsequent light curtain placements.
During training, data instances need to be sampled from the single-beam LiDAR, as well as from up to $K$ number of light curtain placements. We choose $K=3$ in all our experiments. Crucially, since the light curtains are placed based on the output (uncertainty maps) of the detector, the point cloud data distribution from the $k$-th ($1 \leq k \leq K$) light curtain placement depends on the current weights of the detector. As the weights of the detector get updated during each gradient descent step, the input training data distribution from the $k$-th light curtain also changes. To accomodate for non-stationary training data, we propose \textit{training with online data-generation}. This is described in Algorithm~\ref{alg:training}.

\begin{algorithm}
        \caption{Training with Online Light Curtain Data Generation}
        \DontPrintSemicolon
        $W_0 \gets$ initial weights of the detector\;
        $T \gets$ number of training iterations\;
        $K \gets$ number of light curtain placements\;
        \SetKwFunction{FMain}{InputPointCloud}
        \SetKwProg{Fn}{Function}{:}{}
            \Fn{\FMain{$W$, $S$, $k$}}{
                \eIf{$k = 0$} {
                    $P_0 \gets$ point cloud from single-beam LiDAR in scene S\;
                    \KwRet $P_0$\;
                }{
                    $P_{k-1} \gets$ \texttt{InputPointCloud($W$, $S$, $k-1$)}\;
                    $H \gets$ uncertainty map from detector with weights $W$ and input $P_{k-1}$\;
                    $P \gets$ point cloud from placing light curtain optimized for $H$ in scene $S$\;
                    $P_k \gets P_{k-1} \cup P$\;
                    \KwRet $P_k$\;
                }
            }
            \For{t = 1 \text{to} T}{
                $S_t \gets$ $t$-th training scene\;
                $k_t \gets$ randomly sample from $\{0, 1, \dots, K\}$\;
                $P_t \gets$ \texttt{InputPointCloud}($W_{t-1}$, $S_t$, $k_t$)\;
                $W_t \gets$ gradient descent update using previous weights $W_{t-1}$ and input $P_t$\;
            }
            \KwRet $W_T$
        \label{alg:training}
    \end{algorithm}

At each training iteration $t$, we retrieve a scene $S_t$ from the training dataset. To create the input point cloud, we choose to either use the single-beam LiDAR data or $k$ light curtain placements ($1 \leq k \leq K$), each of them with equal probability. For generating the $k$-th light curtain data, we start with the single-beam LiDAR point cloud. Then we successively perform a forward pass through the detector network with the current weights to obtain an uncertainty map. We compute the optimal light curtain placement for this map, gather points returned from placing this curtain, and finally, fuse the points back into the input point cloud. This cycle is repeated $k$ times to obtain the input point cloud to train on. Generating light curtain data in such an \emph{online} fashion ensures that the input distribution doesn't diverge from the network weights during the course of training.

\subsection*{Ablation experiment}

 Here, we perform an ablation experiment on the Virtual KITTI dataset, to evaluate the importance of training with online light curtain data generation. We first collect the entire dataset at the beginning, using the initial weights of the network. Then, we freeze this data and train the detector. The results are shown in Table~\ref{table:ablation}. We see that the accuracy on light curtain data (Table~\ref{table:ablation} rows 2-4) decreases substantially to less 2\%, since this data distribution diverges during training. However, the performance on single-beam LiDAR remains relatively same, since the LiDAR data distribution doesn't change. This demonstrates the importance of re-generating the training data online as the weights of the detector change.

\begin{table*}[h!]
  \centering
  \begin{tabular}{?c?c|c|c|c?}
   \Xhline{0.8pt}
   & \multicolumn{4}{c?}{\bf{Virtual KITTI}}\\
   \hline
   & \multicolumn{2}{c|}{3D mAP} & \multicolumn{2}{c?}{BEV mAP}\\
   \hline
   & 0.5 IoU & 0.7 IoU & 0.5 IoU & 0.7 IoU\\
   \Xhline{0.8pt}
   Single Beam Lidar & \bf{37.68} & \bf{18.65} & \bf{38.14} & \bf{30.08}\\
   \hline
   1 Light Curtain & 1.41 & 0.48 & 1.61 & 0.75\\
   \hline
   2 Light Curtains & 0.73 & 0.38 & 1.22 & 0.58\\
   \hline
   3 Light Curtains & 0.68 & 0.36 & 1.13 & 0.54
   \\
   \Xhline{0.8pt}
  \end{tabular}
  \caption{Performance of the detector trained with single-beam LiDAR and up to three light curtains, without online training data generation. The training dataset is collected using the initial weights of the network and is fixed during the remainder of training. The light curtain performance decreases substantially.}
  \label{table:ablation}
\end{table*}
\section{Noise simulations}
\label{app:noise}

In order to simulate noise in the real-world sensor, we add 10\% noise to the light curtain input, for varying number of light curtain placements, on the Virtual KITTI dataset. The results are shown in Table~\ref{table:noise}. The results are comparable to without noise, indicating that our method is robust to noise and is likely to transfer well to real-world data.

\begin{table*}[h!]
  \centering
  \begin{tabular}{?c?c|c|c|c?c|c|c|c?}
   \Xhline{0.8pt}
   & \multicolumn{8}{c?}{\bf{Virtual KITTI}}\\
   \hline
   & \multicolumn{4}{c?}{\bf{Without noise}} & \multicolumn{4}{c?}{\bf{With noise}}\\
   \hline
   & \multicolumn{2}{c|}{3D mAP} & \multicolumn{2}{c?}{BEV mAP} & \multicolumn{2}{c|}{3D mAP} & \multicolumn{2}{c?}{BEV mAP}\\
   \hline
   & 0.5 IoU & 0.7 IoU & 0.5 IoU & 0.7 IoU & 0.5 IoU & 0.7 IoU & 0.5 IoU & 0.7 IoU\\
   \Xhline{0.8pt}
   Single Beam Lidar & 39.91 & 15.49 & 40.77 & 36.54 & 39.03 &	17.13 &	39.93 &	30.26\\
   \hline
   1 Light Curtain & 58.01 & 35.29 & 58.51 & 47.05 & 57.04	& 25.99 &	57.65 &	45.31\\
   \hline
   2 Light Curtains & 60.86 & 37.91 & 61.10 & 49.84 & 59.43	& 30.91 &	59.89 &	46.11\\
   \hline
   3 Light Curtains & \bf{68.52} & \bf{38.47} & \bf{68.82} & \bf{50.53} & \bf{60.02}	& \bf{31.09} &	\bf{66.78} &	\bf{46.39}
   \\
   \Xhline{0.8pt}
  \end{tabular}
  \caption{Performance of detectors trained with single-beam LiDAR and up to three light curtains, with $10\%$ additional noise in the light curtain input. Performance is not significantly lower than without noise.}
  \label{table:noise}
\end{table*}

\section{Efficiency analysis}

In this section, we report the time taken by our method, for varying number of light curtain placements, and for different light curtain placement algorithms, in Table~\ref{table:timing}. The time (in seconds) includes the time taken for all preceding steps. For example, the time for 2 light curtain placements includes the time required for generating the single-beam LiDAR data, computing the optimal first and second light curtain placements, and all intermediate forwarded passes through the detection network while generating uncertainty maps. The time is averaged over 100 independent trials over different scenes, and we report the 95\% confidence intervals.

\begin{table}
    \centering
    \begin{tabular}{?c?c|c|c|c?} 
     \Xhline{0.8pt}
     & \makecell{Single-beam\\LiDAR} & \makecell{One\\ light curtain} & \makecell{Two\\ light curtain} & \makecell{Three\\ light curtain}\\
     \Xhline{0.8pt}
     Random & 0.096 $\pm$ 0.001 & 0.763 $\pm$ 0.008 & 1.441 $\pm$ 0.014 & 2.133 $\pm$ 0.014 \\
     \hline
     Fixed depth - 15m & 0.090 $\pm$ 0.002 & 0.765 $\pm$ 0.008 & 1.412 $\pm$ 0.012 & 2.028 $\pm$ 0.018 \\
     \hline
     Fixed depth - 30m & 0.095 $\pm$ 0.002 & 0.789 $\pm$ 0.005 & 1.474 $\pm$ 0.008 & 2.180 $\pm$ 0.013 \\
     \hline
     Fixed depth - 45m & 0.094 $\pm$ 0.001 & 0.778 $\pm$ 0.003 & 1.475 $\pm$ 0.013 & 2.174 $\pm$ 0.012 \\
     \hline
     \makecell{Greedy Optimization\\(Randomly break ties)} & 0.092 $\pm$ 0.000 & 0.825 $\pm$ 0.014 & 1.547 $\pm$ 0.023 & 2.250 $\pm$ 0.030 \\
     \hline
     \makecell{Greedy Optimization\\(Min laser angle change)} & 0.086 $\pm$ 0.001 & 0.824 $\pm$ 0.010 & 1.543 $\pm$ 0.020 & 2.242 $\pm$ 0.028 \\
     \hline
     \makecell{Frontoparallel +\\Uncertainty} & 0.091 $\pm$ 0.001 & 0.441 $\pm$ 0.003 & 0.807 $\pm$ 0.006 & 1.165 $\pm$ 0.008 \\
     \hline
     Dynamic Programming & 0.097 $\pm$ 0.008 & 0.944 $\pm$ 0.010 & 1.767 $\pm$ 0.015 & 2.600 $\pm$ 0.020 \\
     \Xhline{0.8pt}
    \end{tabular}
    \caption{Time efficiency (in seconds) for varying number of light curtains and different light curtain placement algorithms. Time is averaged over 100 independent trials over different scenes, and we report the 95\% confidence intervals.}
    \label{table:timing}
\end{table}

\begin{figure}
  \centering
  \includegraphics[trim=11 11 22 11,clip,width=0.5\textwidth]{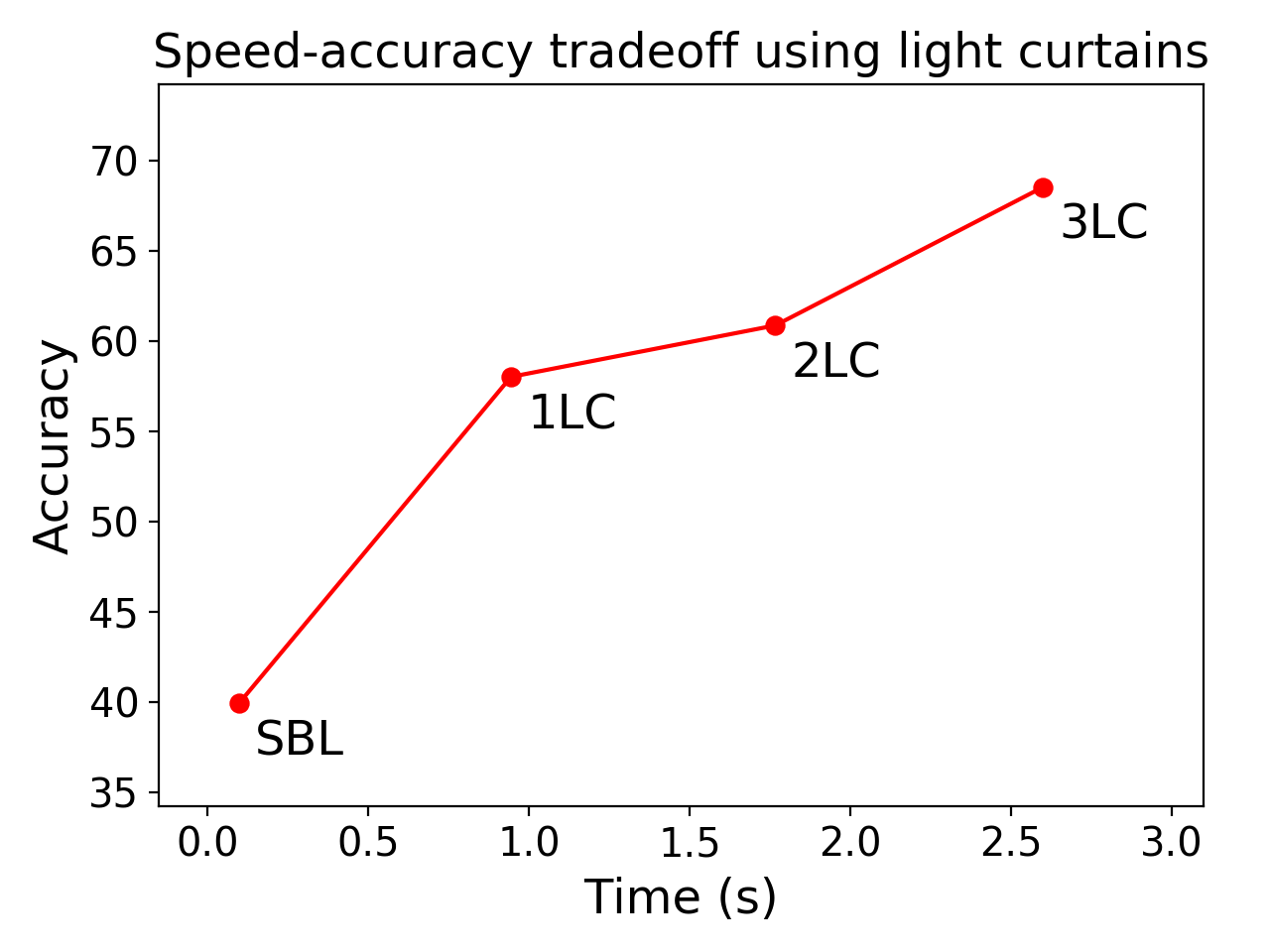}
  \caption{Speed-accuracy tradeoff using light curtains optimized by dynamic programming, on the Virtual KITTI dataset. More light curtains correpsond to increased accuracy but reduced speed.}
  \label{fig:tradeoff}
\end{figure}

Note that as we place more light curtains, more time is consumed for the network's forward pass and in calculating where to place the light curtain. This presents a speed-accuracy tradeoff; more light curtains will improve detection accuracy at the expense of taking more time. On the other hand, our method can run faster using fewer light curtains but with a decreased accuracy. This tradeoff is visualized in Figure~\ref{fig:tradeoff}.

\end{document}